# Systematic Evaluation of Vision Transformers for Automated Cervical Cancer Classification: Optimization, Statistical Validation, and Clinical Interpretability

Nisreen Albzour* and Sarah S. Lam

School of Systems Science and Industrial Engineering,
Binghamton University,
Binghamton, NY 13902, USA

Email: nalbzour@binghamton.edu, sarahlam@binghamton.edu

*Corresponding author: nalbzour@binghamton.edu

**Abstract**
Manual Pap smear analysis for cervical cancer screening is limited by inter-observer variability, time constraints, and restricted expert availability. Although convolutional neural networks (CNNs) have automated cervical cell classification, they remain limited in modeling long-range spatial dependencies and often lack clinical interpretability. In this study, Vision Transformer (ViT) architectures were systematically optimized to enhance automated cervical cancer screening, which resulted in improved interpretability. The Herlev dataset (917 images: 242 normal, 675 abnormal) was utilized to optimize ViT-Tiny, a lightweight Vision Transformer architecture designed for reduced computational complexity, through a comprehensive evaluation of augmentation strategies, class weighting, and hyperparameters. The optimal configuration achieved 94.9%–95.2% cross-validation accuracy, in which random horizontal flipping and class weighting (0.7 × 1.3) were identified as most effective. Gradient-weighted Class Activation Mapping (Grad-CAM) analysis confirmed that model attention corresponded to clinically relevant morphological features, which include nuclear regions, cell boundaries, and chromatin texture, which align with cytopathological criteria. These findings indicate that Vision Transformers can deliver accurate and interpretable decision support for cervical cancer screening, which fulfills both clinical performance and transparency requirements essential for medical AI deployment.

## 1. Introduction

Cervical cancer remains a significant global health challenge, in which over 600,000 new cases and 340,000 deaths are reported annually worldwide [1]. The primary strategy for reducing mortality is early detection through Pap smear screening, which enables identification of precancerous lesions before they progress to invasive cancer. However, manual analysis of Pap smear slides is inherently time-consuming, labor-intensive, and subject to substantial interobserver variability, in which diagnostic accuracy varies significantly among cytotechnologists [2-3]. These limitations highlight the critical need for automated, reliable, and interpretable diagnostic solutions that can augment human expertise while maintaining the accuracy essential for clinical decision-making.

Deep learning approaches, particularly Convolutional Neural Networks (CNNs), have revolutionized medical image analysis by providing automated feature extraction and achieving remarkable diagnostic accuracy across various medical imaging modalities [4-5]. CNNs have demonstrated substantial success in cervical cancer detection, which surpass traditional machine learning methods through their ability to learn hierarchical feature representations directly from image data[7-8]. However, CNN architecture faces inherent limitations in medical imaging applications. Their reliance on local receptive fields and hierarchical feature extraction can miss important long-range spatial relationships that may be crucial for accurate diagnosis of cellular abnormalities[9-10]. Additionally, the limited interpretability of CNN decision-making processes poses significant barriers to clinical adoption, where understanding the rationale behind diagnostic predictions is essential for building trust and enabling clinical validation [11-12].
Related automation-oriented work in healthcare has also explored hyperautomation-based leukemia detection and classification [6], supporting the broader relevance of AI-driven clinical decision-support workflows beyond a single disease domain.

Vision Transformers (ViTs) have emerged as a promising alternative architecture that addresses many limitations of traditional CNNs. By adapting the transformer architecture from natural language processing to computer vision, ViTs [13] leverage self-attention mechanisms to capture global contextual relationships across entire images, which enable more comprehensive analysis of spatial patterns [14]. This global perspective is particularly valuable for medical image analysis, where diagnostic features may be distributed across different regions of an image and require integration of multiple cellular characteristics [15]. Furthermore, the attention mechanisms inherent in ViTs provide natural interpretability through attention maps, which offer insights into which image regions influence diagnostic decisions—a crucial requirement for clinical acceptance and regulatory approval [16]. Recent studies have further validated the effectiveness of Vision Transformers in medical imaging applications, demonstrating improved performance in modeling both global and local context, enhanced computational efficiency, and robustness to limited labeled data [17-18].

Despite the promising potential of ViTs in medical imaging, their application to cervical cancer screening remains underexplored. Existing studies have not systematically examined optimization requirements for ViTs in this domain, nor have they provided the rigorous statistical validation necessary for clinical adoption. Critical gaps remain in the evaluation of data augmentation strategies, class imbalance handling techniques, and hyperparameter optimization tailored for cervical cell classification. Furthermore, the interpretability advantages of ViTs have not been fully investigated in the context of cytopathological analysis, where alignment with established diagnostic criteria is essential.

This study addresses these gaps by presenting a comprehensive evaluation of ViT architectures for automated cervical cancer screening using Pap smear images. The contributions of this study are fourfold: (1) systematic optimization of augmentation strategies and class weighting approaches to address class imbalance while preserving biological validity; (2) rigorous statistical validation through repeated experiments and pairwise comparisons to identify statistically equivalent high-performing configurations; (3) enhanced Gradient-weighted Class Activation Mapping (Grad-CAM) interpretability aligned with cytopathological diagnostic criteria; and (4) demonstration that ViTs can achieve clinically relevant performance while providing the transparency and reliability essential for medical AI applications.

The remainder of this paper is organized as follows: Section 2 reviews prior work on traditional machine learning, deep learning, and emerging transformer-based methods for cervical cancer detection. Section 3 provides a detailed description of the methodology, which includes dataset preparation, model architecture, optimization strategies, and interpretability frameworks. Section 4 presents experimental results and discussion, which include statistical analyses and interpretability findings. Section 5 concludes with a summary of contributions and directions for future research.

## 2. Related Literature

The application of artificial intelligence techniques in cervical cancer detection and classification has advanced significantly over the past decade, which led to the development of increasingly accurate and automated diagnostic methods. As illustrated in Figure 1, these techniques can be broadly categorized into machine learning and deep learning approaches, in which each category encompasses a range of methodologies tailored to improve diagnostic performance.

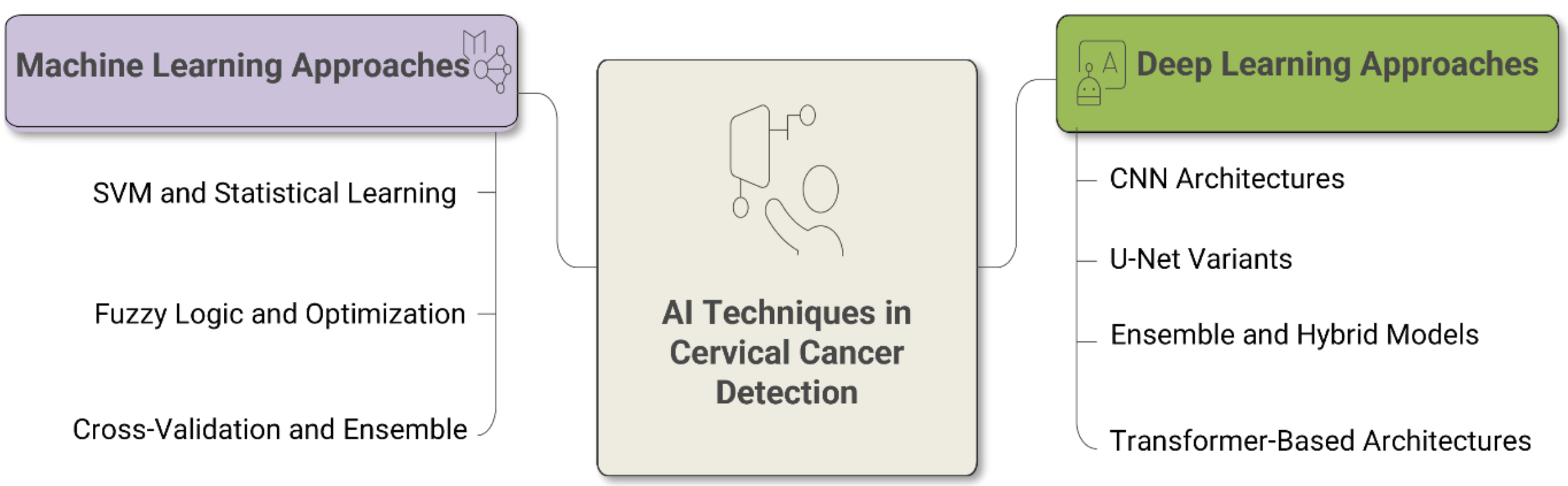

*Figure 1: AI Techniques in Cervical Cancer Detection*

### 2.1 Machine Learning Approaches

Traditional machine learning approaches for cervical cancer detection rely heavily on handcrafted feature extraction followed by classical classification algorithms. This subsection reviews studies that employ various feature-engineering and machine-learning techniques. Support Vector Machine (SVM) and statistical learning approaches have been extensively utilized. Several studies [19-21] developed hybrid linear iterative clustering with Bayes classification, k-Nearest Neighbors (k-NN) with fuzzy logic, and machine learning-assisted detection frameworks, respectively. Mariarputham and Stephen [22] used handcrafted texture features with SVM and neural networks on the Herlev dataset, which found SVM performed best. Optimization-based methods [23] explored quantum hybrid Particle Swarm Optimization (PSO) with fuzzy k-NN for feature selection and cell classification. Cross-validation and prediction-based machine-learning techniques [24-26] investigated stratified k-fold cross-validation, early detection frameworks, and prediction using various machine learning methods. Additional ML approaches addressed Pap smear image classification and early prediction [27-28]. Nandanwar and Dhonde [29] proposed a hybrid stacked ensemble model that combines feature extraction with segmentation, followed by classifier-level fusion using Histogram and Hu Moments techniques.

Traditional machine learning approaches in cervical cancer diagnosis often lack integrated data augmentation strategies, which limits their ability to generalize across diverse image variations. Furthermore, these systems are prone to high false negative rates, which can result in missed diagnoses and pose significant risks in clinical settings. These limitations highlight the need for more robust, automated, and generalizable solutions in medical diagnostics.

### 2.2 Deep Learning Approaches

Deep learning approaches have demonstrated remarkable success in medical image analysis, particularly for cervical cancer detection and classification. The following review encompasses studies that employ various deep neural network architectures for cervical cancer classification.

CNNs are widely adopted deep learning models known for their ability to extract spatial and hierarchical features from input data, which make them suitable for a variety of pattern recognition tasks. CNN architecture has been extensively employed for cervical cancer detection across multiple imaging modalities. Several studies [30-33] employed various CNN variants that included Mask Region-based Convolutional Neural Networks (R-CNN) with Visual Geometry Group (VGG) and Residual Network (ResNet) components, Capsule Networks (CapsNet) preprocessing with VGG-like networks, specialized CNN architectures using deformable Faster Region-based Convolutional Neural Network with Feature Pyramid Networks (R-CNN-FPN) with pretrained backbones for cervical image analysis, and improved Inception-ResNet-V2. Transfer learning approaches [34] implemented and optimized Squeeze-and-Excitation Residual Network-152 (SE-ResNet152) models for multi-class cervical cancer detection. Recent CNN-based work [35-38] explored privacy-preserved detection, lightweight CNN architectures, ResNet-based automated screening, and Attention-Fused Squeeze-and-Excitation Network (AF-SENet) using pre-trained models for feature fusion. Additional CNN approaches [39-42] investigated various CNN variants with different backbone networks for enhanced performance.
In a closely related cervical cytology study, Albzour and Lam investigated deep-learning-based segmentation and classification of Pap smear images for cervical cancer detection [43], which provides a relevant foundation for the present transformer-based extension.

U-Net is a deep learning architecture originally developed for biomedical applications, which is recognized for its encoder-decoder structure and ability to capture both high-level context and fine-grained details. U-Net variants and advanced architectures have shown exceptional performance in cervical cancer applications. Enhanced U-Net architectures[44-46] incorporated residual SE blocks, feature fusion from multiple pretrained models, and encoder-weighted designs for improved analysis.

Ensemble learning refers to the technique of combining predictions from multiple models to achieve better performance and generalization than any single model alone. Ensemble and hybrid deep learning approaches have emerged to improve robustness and diagnostic accuracy in cervical cancer classification. Several studies have proposed hybrid frameworks that integrate deep feature fusion, pretrained networks, and classifier-level combinations to enhance performance[47-49] . In addition, radiomics-guided deep learning approaches have combined handcrafted feature descriptors with neural network representations to further strengthen feature expressiveness and diagnostic robustness [50].

Following advancements in CNNs, U-Nets, and ensemble learning, transformer-based architectures have recently attracted significant attention due to their capability to capture global contextual relationships [51]. Recent studies have further advanced Vision Transformer frameworks through systematic optimization strategies and attention-enhanced architectures to improve classification accuracy, computational efficiency, and contextual feature learning in medical image analysis [52-54]. Vision Transformers have demonstrated strong potential in learning long-range dependencies in medical images, though their performance often benefits from pretraining and large datasets.

Despite their successes, deep learning approaches face several limitations that impact their clinical applicability. Common challenges include class imbalance between normal and abnormal cells, and lack of interpretability tools such as saliency maps or attention visualizations. These issues can reduce model robustness and make predictions less transparent and more difficult to validate for clinical use. This study specifically addresses two critical limitations: class imbalance through systematic optimization of class weighting approaches, and lack of interpretability through comprehensive Grad-CAM visualization that aligns model attention with established cytopathological criteria. Addressing these limitations is essential to improve trust, generalizability, and the practical deployment of AI models in cervical cancer diagnosis.

## 3. Methodology

Figure 2 provides a comprehensive overview of the entire methodology employed in this study. Each stage plays a crucial role that ensures the effectiveness and robustness of the classification framework, which underscores the systematic approach adopted to tackle the challenges in cervical cancer diagnosis.

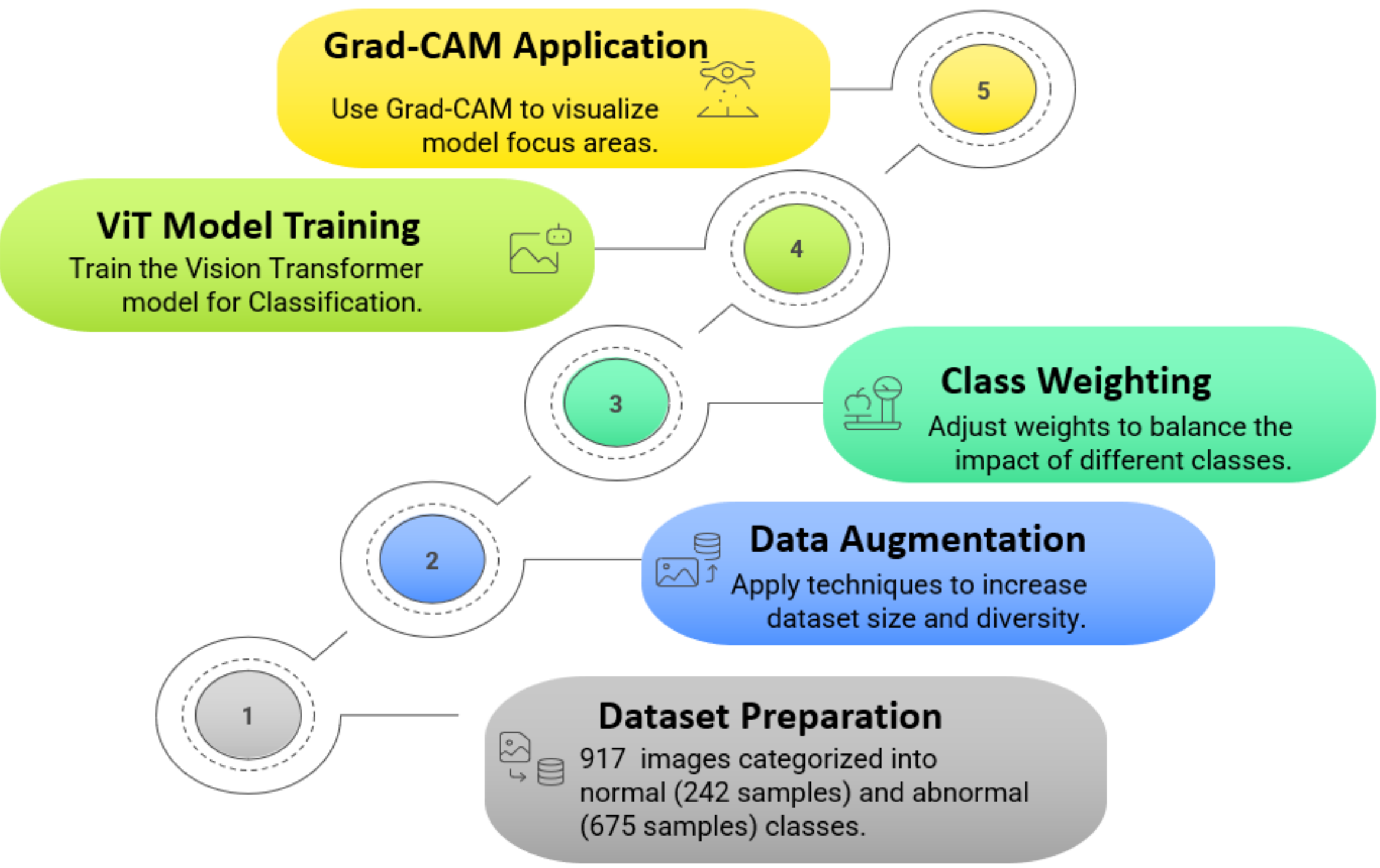

*Figure 2: Overview of the Methodology*

### 3.1 Dataset Preparation

The Herlev Pap Smear dataset, which comprises 917 images, was used and regrouped from seven cytological classes into a binary screening task (normal: 242; abnormal: 675), consistent with clinical triage. Images were inspected for readability and converted to a unified input size compatible with ViT (RGB, standardized resolution). Pixel intensities were normalized using ImageNet statistics to leverage pretrained foundations. To ensure unbiased assessment, stratified k-fold cross-validation with patient-level separation was employed, which ensures that no image from the same subject appears in both training and validation folds. All preprocessing and augmentations were applied on the fly within the training pipeline to avoid data leakage.

### 3.2 Data Augmentation

Augmentations were designed to increase data diversity while preserving cytological validity (cell/nuclear morphology, chromatin patterns). The augmentation policy included the following transformations:

- Geometric: horizontal flip, small rotations, translations, and mild scaling to simulate slide handling variability.
- Photometric: limited brightness/contrast jitter and color perturbations to reflect staining variability.
- Regularization: light Gaussian blur/noise where appropriate. Each transformation used clinically conservative ranges and per-operation probabilities to ensure that augmented images remained biologically plausible. Augmentations were applied only to training data.

### 3.3 Class Weighting

Given the class imbalance (normal < abnormal), class-weighted cross-entropy was employed to mitigate bias toward the majority class. For class $c$ with $N_c$ samples and a total of $N$ samples across $C$ classes, the class weight $W_c$ was defined as:

$$W_c = \frac{N}{C \times N_c}$$

where $N$ is the total number of samples, $C$ is the total number of classes, and $N_c$ is the number of samples in class $c$. For the binary case ($C = 2$), this formulation assigns a larger weight to the minority (normal) class, thereby penalizing its misclassification more heavily.

### 3.4 ViT Model Training

The primary backbone was a Vision Transformer (ViT) adapted for binary classification by replacing the classification head with a two-logit linear layer. The encoder was initialized with ImageNet pretraining to enhance data efficiency. Training used a modern optimization recipe: AdamW with weight decay, label smoothing, and a learning-rate schedule (warm-up followed by cosine/step decay). Early stopping and checkpointing were triggered by validation loss to prevent overfitting. Hyperparameters (learning rate, batch size, epochs, weight decay, label smoothing) were explored within standard ranges using grid/random search; the selection criteria were validation metrics defined in Section 3.6.

### 3.5 Grad-CAM Application

To enhance model interpretability, Grad-CAM was applied to ViT predictions using a transformer-compatible implementation that computes gradients with respect to the final attention or feature maps. Heatmaps were generated for validation images to visualize regions that contribute to the classification decision. Visualization parameters (smoothing, normalization, overlay opacity) were maintained consistently across all runs. This procedure enables qualitative assessment of whether the model's attention corresponds to clinically relevant morphological regions such as nuclei, irregular boundaries, or chromatin distribution. The resulting observations are analyzed in Section 4.

### 3.6 Evaluation Protocol

For completeness, performance was assessed with accuracy, precision, recall, and F1-score on the held-out folds. The metrics are defined as follows:

$$Accuracy = \frac{TP + TN}{TP + TN + FP + FN}$$

$$Precision = \frac{TP}{TP + FP}$$

$$Recall\ (Sensitivity) = \frac{TP}{TP + FN}$$

$$F1 - score = \frac{2 \cdot (Precision \cdot Recall)}{Precision + Recall}$$

where $TP$, $TN$, $FP$, and $FN$ represent true positives, true negatives, false positives, and false negatives, respectively. To enable consistent comparison across configurations, fold-wise means and 95% confidence intervals (CI) were computed as follows:
The use of multiple evaluation metrics and feature-based model assessment is also consistent with prior health analytics research on post-stroke activities of daily living prediction, where feature selection was used to improve machine learning performance and interpretability [55].

$$CI = \bar{x} \pm z \cdot \frac{s}{\sqrt{n}}$$

where $\bar{x}$ = mean of fold-wise results, $s$ = standard deviation, $n$ = number of folds, $z = 1.96$ (for 95% confidence interval). Interpretability outputs were summarized with representative heatmaps; quantitative alignment with expert-defined regions can be added when annotations are available.

## 4. Results and Discussion

This section reports experimental outcomes following the same five-stage workflow described in Section 3 and Figure 2: (1) Dataset Preparation, (2) Data Augmentation, (3) Class Weighting, (4) ViT Model Training, and (5) Grad-CAM Interpretability. Organizing results in parallel with methodology enables direct tracing from design choices to performance.

### 4.1 Dataset Preparation Results

The Herlev dataset was successfully preprocessed into 917 images with binary classification (242 normal, 675 abnormal). All images were standardized to RGB format with consistent resolution. ImageNet normalization statistics were applied (mean= [0.485, 0.456, 0.406], std= [0.229, 0.224, 0.225]). Stratified 5-fold cross-validation maintained the 27.8% normal, 72.2% abnormal distribution across all folds.

### 4.2 Data Augmentation Results

Table 1 presents the performance comparison of seven augmentation strategies (three single and four combined augmentation strategies) evaluated using 5-fold cross-validation with optimal class weights.

*Table 1: Augmentation Techniques Evaluation (5-Fold Cross-Validation)*

| Augmentation Strategy | Precision (%) | Recall (%) | F1-score | Accuracy (%) |
|---|---|---|---|---|
| **Color Jitter** | 80.40 | 90.60 | 83.30 | 89.87 |
| **Horizontal Flip** | **89.70** | **91.30** | **90.00** | **94.77** |
| **Random Affine** | 86.90 | 83.40 | 83.10 | 91.39 |
| **Color Jitter + Horizontal Flip** | 89.10 | 90.10 | 89.40 | 94.33 |
| **Color Jitter + Random Affine** | 84.10 | 95.00 | 88.70 | 93.23 |
| **Horizontal Flip + Random Affine** | 83.60 | 89.70 | 85.70 | 92.26 |
| **All Three Combined** | 89.40 | 89.70 | 89.10 | 94.22 |

Single augmentations showed varying effectiveness. Color Jitter performed worst with an accuracy of 89.87% and a precision of 80.40%, which suggests color variations obscured key diagnostic cues such as chromatin texture. Random Affine resulted in an accuracy of 91.39% but with a reduced recall (83.40%), likely due to geometric distortions that affected nuclear shape. In contrast, Horizontal Flip achieved strong overall performance 94.77% accuracy and 91.30% recall, which indicates that left–right invariance is a beneficial augmentation for cervical cytology images without distorting diagnostically relevant structures.

Combined augmentations showed limited synergy. Color Jitter + Horizontal Flip (94.33%) was slightly below horizontal flip alone, while Color Jitter + Random Affine (93.23%) achieved the highest recall (95.00%) but lower precision (84.10%), which increased false positives. Horizontal Flip + Random Affine (92.26%) and All Three Combined (94.22%) offered no significant gains over simpler (single) augmentations.

Overall, augmentation effectiveness depends on biological plausibility rather than variety. Transformations that preserve diagnostic features (horizontal flip) outperformed those introducing artificial distortions, which confirms that simpler (single) augmentations as well as clinically valid augmentations yield the most reliable results.

### 4.3 Class Weighting Results

Table 2 summarizes the systematic evaluation of five weight multiplier configurations to address class imbalance.

*Table 2: Class Weight Optimization Results (5-Fold Cross-Validation)*

| Case # | Weight Multiplier | Abnormal Weight | Normal Weight | Precision (%) | Recall (%) | F1-score (%) | Accuracy (%) |
|---|---|---|---|---|---|---|---|
| 1 | 1.0×1.0 | 0.68 | 1.90 | 92.10 | 84.30 | 87.40 | 93.67 |
| 2 | 0.8×0.8 | 0.54 | 1.52 | 84.40 | 90.60 | 85.80 | 91.93 |
| 3 | 1.2×1.2 | 0.82 | 2.27 | 83.00 | 93.00 | 86.70 | 91.72 |
| 4 | 0.7×1.3 | 0.48 | 2.46 | 90.90 | 93.40 | 91.90 | 95.64 |
| 5 | 1.3×0.7 | 0.88 | 1.33 | 90.70 | 88.80 | 89.70 | 94.55 |

Varying the abnormal-to-normal weight ratio produced distinct performance trade-offs. Case 4 (0.7 × 1.3 multipliers) achieved a balanced trade-off between recall and precision, which resulted in the highest F1-score (91.90%). Case 4 achieved the best overall performance, with 95.64% accuracy and 93.40% recall, which effectively prioritizes abnormal-cell detection without excessive misclassification of normal samples. These results confirm that moderate weighting enhances screening reliability, which is critical for medical applications where missing abnormal cases carries higher clinical risk.

### 4.4 ViT Model Training Results

A comprehensive hyperparameter optimization study evaluated 27 combinations of batch sizes (16, 32, 64), learning rates (0.0001, 0.0005, 0.001), and numbers of epochs (5, 10, 15). All hyperparameter configurations were evaluated using stratified 5-fold cross-validation, and Table 3 reports the average performance across folds for the top-performing configurations.

*Table 3: Selected Hyperparameter Optimization Results*

| Experiment # | Batch Size | Learning Rate | Epochs | Precision (%) | Recall (%) | F1-score (%) | Accuracy (%) |
|---|---|---|---|---|---|---|---|
| 3 | 16 | 0.0001 | 15 | 93.36 | 90.52 | 91.82 | 95.75 |
| 11 | 32 | 0.0001 | 10 | 90.47 | 95.06 | 92.59 | 95.96 |
| 12 | 32 | 0.0001 | 15 | 96.23 | 90.49 | 93.10 | 96.51 |
| 21 | 64 | 0.0001 | 15 | 93.83 | 92.15 | 92.87 | 96.29 |

A learning rate of 0.0001 consistently outperformed higher values, which ensured stable convergence and prevented accuracy loss caused by overshooting the optimal loss minimum during optimization. Batch size affected training efficiency; Batch 16 required more epochs due to noisy gradients, while Batch 32 achieved the best balance with the highest accuracy (96.51%). Batch 64 showed slight degradation, likely from reduced gradient diversity. Training duration analysis indicated that 15 epochs yielded optimal convergence, whereas shorter runs led to underfitting.

To assess result stability and reduce the impact of stochastic training effects, each selected configuration was replicated 10 times using different random initializations. Replication mitigates variability introduced by random weight initialization and data shuffling, which provides a more reliable estimate of model performance. Table 4 reports the mean and standard deviation across these replications for both cross-validation and application-level evaluation.

*Table 4:Comprehensive Experimental Results (10 Replications)*

| **Configuration (B = Batch size, E = Epochs)** | **CV Precision (%)** | **CV Recall (%)** | **CV F1-score (%)** | **CV Accuracy (%)** | **App Precision (%)** | **App Recall (%)** | **App F1-score (%)** | **App Accuracy (%)** |
|---|---|---|---|---|---|---|---|---|
| B16_E15 | 91.08 ± 4.10 | 90.99 ± 1.43 | 90.74 ± 1.93 | 95.05 ± 1.20 | 96.98 ± 4.15 | 96.65 ± 6.85 | 96.63 ± 4.27 | 98.27 ± 2.10 |
| B32_E10 | 86.91 ± 3.86 | 93.02 ± 2.82 | 89.36 ± 2.06 | 93.93 ± 1.41 | 97.59 ± 2.48 | 99.55 ± 0.47 | 98.54 ± 1.32 | 99.21 ± 0.72 |
| B32_E15 | 90.78 ± 2.41 | 91.91 ± 1.49 | 91.03 ± 0.87 | 95.15 ± 0.57 | 93.57 ± 14.15 | 99.96 ± 0.12 | 96.00 ± 9.02 | 97.18 ± 6.67 |
| B64_E15 | 91.22 ± 3.48 | 90.38 ± 1.94 | 90.53 ± 1.91 | 94.92 ± 1.22 | 98.41 ± 2.71 | 99.55 ± 0.84 | 98.95 ± 1.34 | 99.43 ± 0.73 |

As shown in Table 4, B32_E15 (batch size of 32 and 15 epochs) achieved the highest cross-validation accuracy (95.15%) with minimal variation, which indicates stable convergence and strong generalization. While all four configurations performed well, application results consistently exceeded cross-validation scores, which is not uncommon because those results are calculated by resubstituting the entire set of data in the model that was constructed using all available data.

Statistical analysis that used pairwise t-tests (Tables 5 and 6) verified if the performance differences in both accuracy and F1-score were significant. The statistical analysis in Tables 5 and 6 revealed that only the comparison between B32_E10 and B32_E15 showed statistically significant differences ($p = 0.035$ for accuracy, $p = 0.045$ for F1-score). The remaining configurations (B16_E15, B32_E15, and B64_E15) exhibited no significant differences, which indicate comparable performance levels. These results confirm that model variations were minor beyond the optimal configuration, as visualized in Figure 3.

*Table 5: Pairwise T-Test Results for Accuracy*

| **Comparison** | **Mean₁ (Exp A)** | **Mean₂ (Exp B)** | **Diff (A-B)** | **95% CI (Diff)** | **p-value** | **Significant? ($p < 0.05$)** |
|---|---|---|---|---|---|---|
| **Exp1 vs Exp2** | 95.05 | 93.93 | +1.122 | (−0.179, 2.423) | 0.087 | ❌ No |
| **Exp1 vs Exp3** | 95.05 | 95.15 | −0.099 | (−1.064, 0.865) | 0.826 | ❌ No |
| **Exp1 vs Exp4** | 95.05 | 94.92 | +0.128 | (−1.077, 1.334) | 0.825 | ❌ No |
| **Exp2 vs Exp3** | **93.93** | **95.15** | **−1.221** | **(−2.336, −0.106)** | **0.035** | ✅ **Yes** |
| **Exp2 vs Exp4** | 93.93 | 94.92 | −0.993 | (−2.306, 0.319) | 0.129 | ❌ No |
| **Exp3 vs Exp4** | 95.15 | 94.92 | +0.228 | (−0.753, 1.209) | 0.622 | ❌ No |

*Table 6: Pairwise T-Test Results for F1- Score*

| **Comparison** | **Mean₁ (Exp A)** | **Mean₂ (Exp B)** | **Diff (A-B)** | **95% CI (Diff)** | **p-value** | **Significant? ($p < 0.05$)** |
|---|---|---|---|---|---|---|
| **Exp1 vs Exp2** | 90.74 | 89.36 | +1.384 | (−0.603, 3.371) | 0.160 | ❌ No |
| **Exp1 vs Exp3** | 90.74 | 91.03 | −0.287 | (−1.824, 1.250) | 0.692 | ❌ No |
| **Exp1 vs Exp4** | 90.74 | 90.53 | +0.212 | (−1.697, 2.121) | 0.817 | ❌ No |
| **Exp2 vs Exp3** | **89.36** | **91.03** | **−1.671** | **(−3.297, −0.045)** | **0.045** | ✅ **Yes** |
| **Exp2 vs Exp4** | 89.36 | 90.53 | −1.172 | (−3.149, 0.805) | 0.228 | ❌ No |
| **Exp3 vs Exp4** | 91.03 | 90.53 | +0.499 | (−1.024, 2.021) | 0.489 | ❌ No |

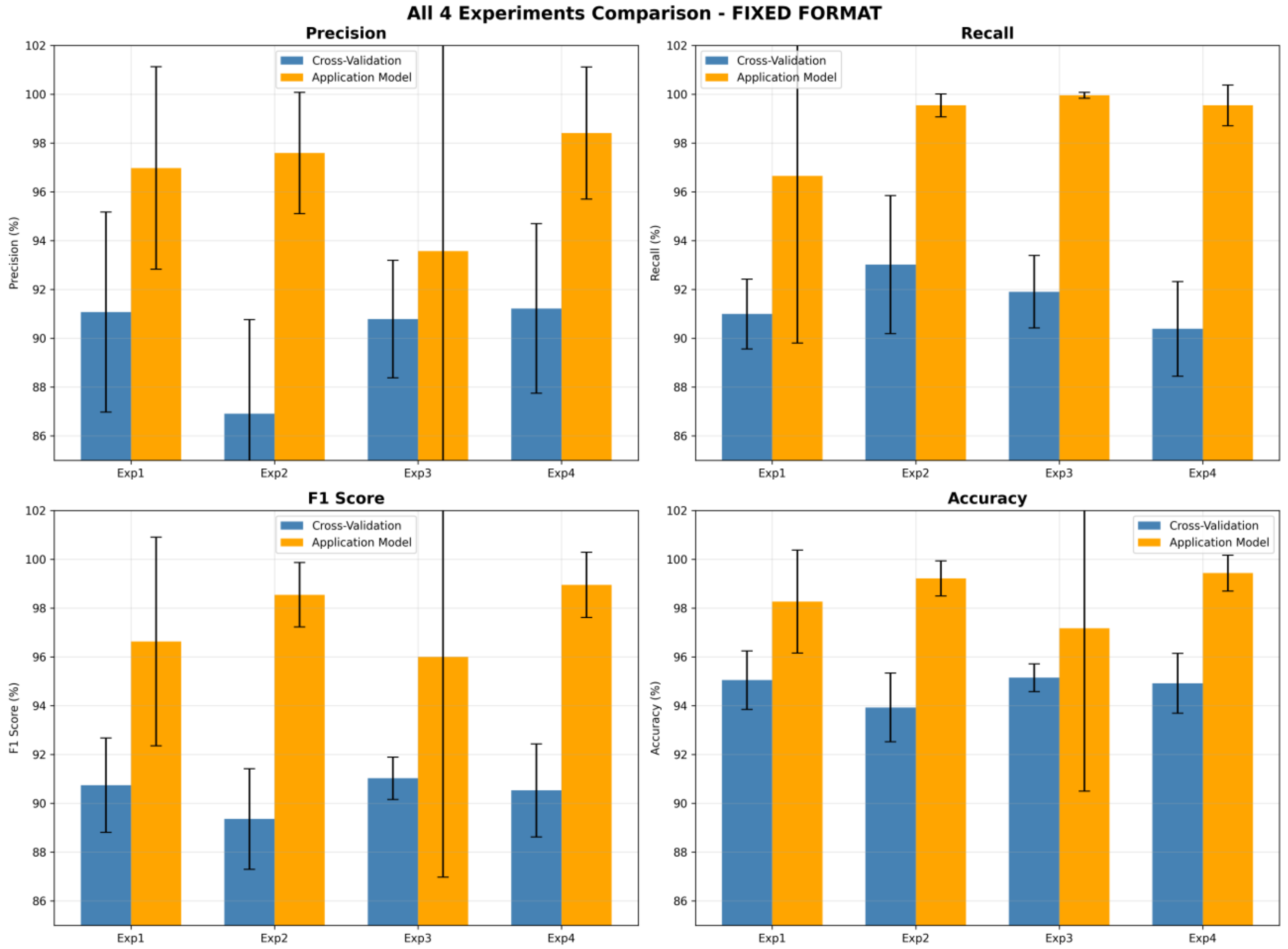

*Figure 3: Performance Comparison Across All Experimental Configurations*

### 4.5 Grad-CAM Interpretability Results

Gradient-weighted Class Activation Mapping was applied to the best-performing configuration (B32_E15) to visualize model decision-making patterns and provide transparency in the classification process.

Figure 4 presents Grad-CAM visualizations for correctly classified abnormal cells and false negative errors. For correctly classified abnormal cells, the model demonstrated strong focus on nuclear regions (attention score: 1.000) with minimal attention to normal features (score: 0.000). In contrast, false negative cases shown in Figure 4 exhibited high normal focus scores (0.994-0.998) with insufficient attention to abnormal features (0.002-0.006), which indicates the model incorrectly focused on normal-appearing regions within abnormal samples.

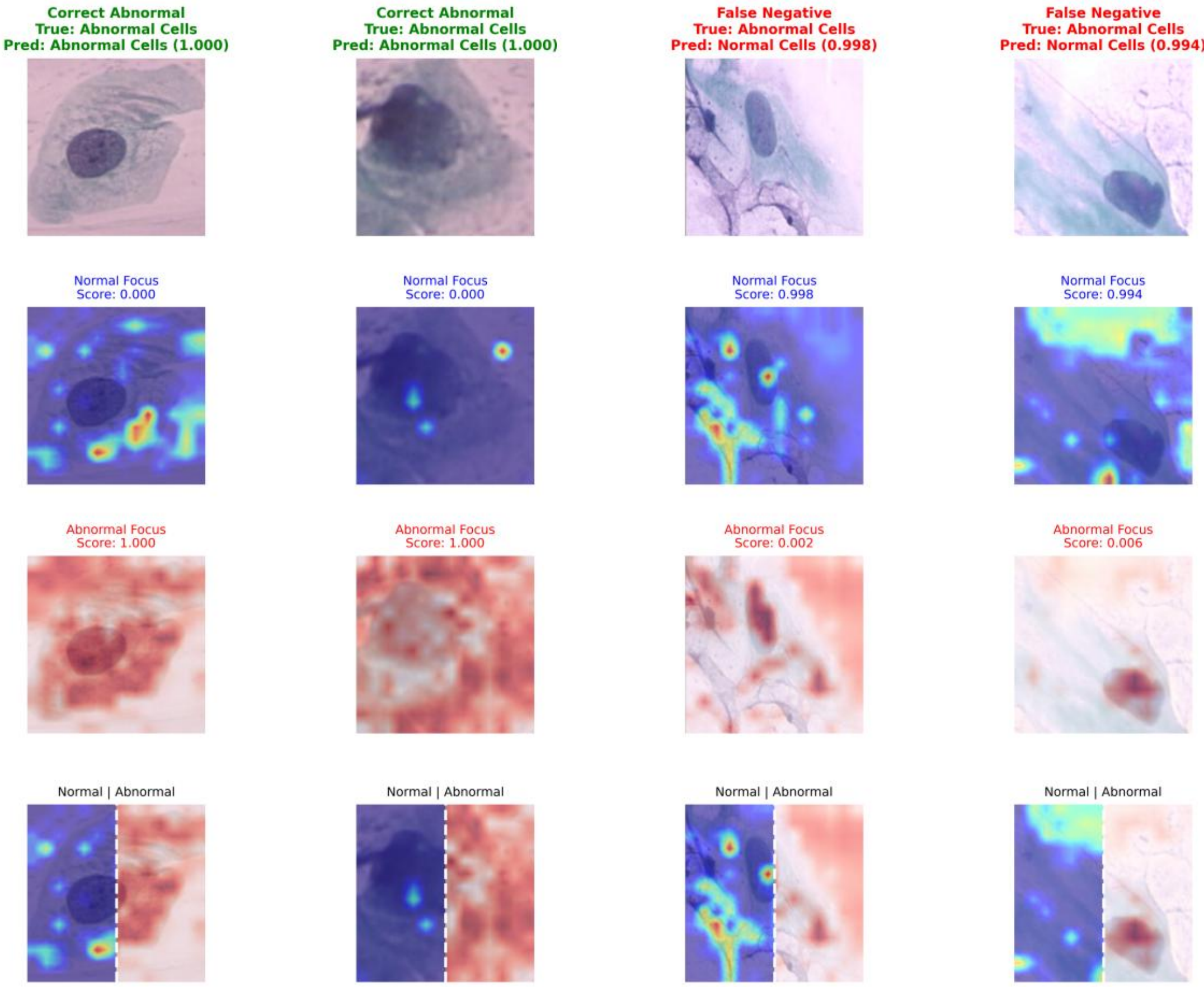

*Figure 4: Grad-CAM Analysis of Correct Abnormal Classifications and False Negative Errors*

Figure 5 illustrates the attention patterns for false positive errors and correctly classified normal cells. False positive cases in Figure 5 showed moderate abnormal focus scores (0.712-0.725) on benign features that superficially resembled abnormal patterns, which often result from staining artifacts or cellular overlapping. Correctly classified normal cells demonstrated appropriate normal focus (0.997-1.000) with minimal abnormal attention (0.000-0.003).

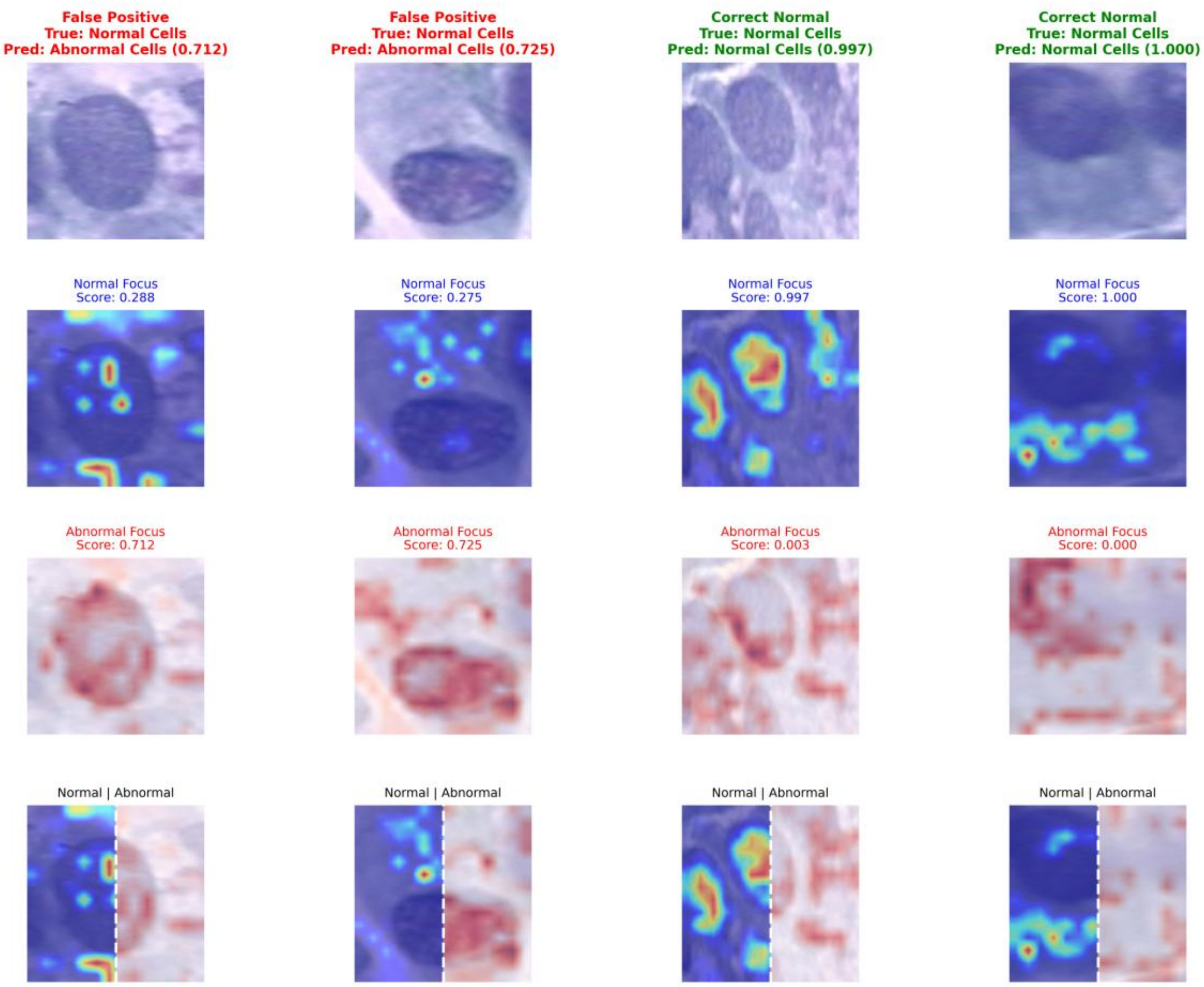

*Figure 5: Grad-CAM Analysis of False Positive Errors and Correct Normal Classifications*

The attention patterns revealed in Figures 4 and 5 consistently aligned with established cytopathological criteria, which focus on nuclear morphology, chromatin distribution, and cellular boundaries—the same features utilized by trained cytopathologists for diagnosis. This alignment demonstrates that the Vision Transformer model learns clinically relevant features for cervical cancer detection.

## 5. Conclusion

This study demonstrates the potential of Vision Transformers as accurate and interpretable tools for automated cervical cancer screening. By restructuring Pap smear classification into a binary screening task, the experiments were aligned with clinical practice and demonstrated that optimized ViT models can achieve ~95% accuracy with statistically validated performance across multiple configurations. Interpretability analysis confirmed that the models consistently focused on clinically meaningful morphological features, which reinforce their potential as trustworthy decision-support tools.

Several limitations should be acknowledged in this study. First, the Herlev dataset contains only 917 images from a single institution, which may not fully represent the diversity of cervical cell morphology across different populations, staining protocols, and imaging equipment. Second, the binary classification approach, while clinically relevant for initial screening, simplifies the nuanced multi-class grading system used in cytopathology practice. Third, the study relied on pre-existing annotations without inter-rater reliability assessment, which potentially introduce label noise. Fourth, the computational requirements of Vision Transformers may limit deployment in resource-constrained settings where cervical cancer screening is most needed. Finally, while Grad-CAM provides valuable insights into model attention, it may not fully capture all aspects of the transformer's decision-making process, particularly the complex interactions between attention heads.

While these results are promising, further validation on larger and multi-center datasets is essential to ensure robustness across diverse populations and imaging protocols. Additionally, prospective clinical trials that evaluate AI-assisted screening in real-world workflows with cytotechnologists and pathologists are crucial for clinical translation, which assess not only diagnostic accuracy but also workflow integration, efficiency, and user acceptance in routine medical practice. Ultimately, this research contributes an important step toward bridging cutting-edge AI with clinical practice, which paves the way for AI-assisted cytopathology systems that can enhance early cervical cancer screening, reduce diagnostic errors, and expand global access to preventive care.